\documentclass[a4paper, 10pt, conference]{ieeeconf}      
\usepackage{FG2025}

\FGfinalcopy 

\usepackage{graphicx}
\usepackage{amsmath}
\usepackage{mathtools} 
\usepackage{amssymb}
\usepackage{times}
\usepackage{epsfig}
\usepackage{caption}
\usepackage{subcaption}
\usepackage{comment}
\usepackage{multirow}
\usepackage{paralist}
\usepackage{booktabs}
\definecolor{redarrow}{RGB}{208,1,1}
\definecolor{darkgreen}{RGB}{0,176,80}

\IEEEoverridecommandlockouts                              
\def\FGPaperID{247} 

\makeatletter
\let\NAT@parse\undefined
\makeatother
\usepackage[colorlinks]{hyperref}
\hypersetup{
    linkcolor=blue,
    filecolor=magenta,      
    urlcolor=cyan,
    citecolor=green    
    }

\title{\LARGE \bf
2D-3D Attention and Entropy for Pose Robust 2D Facial Recognition
}

\author{\parbox{16cm}{\centering 
    {\large J. Brennan Peace$^1$, Shuowen Hu$^2$, and Benjamin S. Riggan$^1$}\\
    {\normalsize
    $^1$ IMAGES Lab., ECE Dept., University of Nebraska-Lincoln, Lincoln, Nebraska, USA\\
    $^2$ U.S. Army Research Laboratory, Adelphi, Maryland, USA}}
}

\usepackage{fancyhdr}
\begin{document}

\ifFGfinal
\thispagestyle{empty}
\pagestyle{empty}
\else
\author{Anonymous FG2025 submission\\ Paper ID \FGPaperID \\}
\pagestyle{plain}
\fi
\maketitle

\thispagestyle{fancy}
 
\begin{abstract}
  Despite recent advances in facial recognition, there remains a fundamental issue concerning degradations in performance due to substantial perspective (pose) differences between enrollment and query (probe) imagery. Therefore, we propose a novel domain adaptive framework to facilitate improved performances across large discrepancies in pose by enabling image-based (2D) representations to infer properties of inherently pose invariant point cloud (3D) representations. Specifically, our proposed framework achieves better pose invariance by using (1) a shared (joint) attention mapping to emphasize common patterns that are most correlated between 2D facial images and 3D facial data and (2) a joint entropy regularizing loss to promote better consistency---enhancing correlations among the intersecting 2D and 3D representations---by leveraging both attention maps. This framework is evaluated on FaceScape and ARL-VTF datasets, where it outperforms competitive methods by achieving profile ($90^\circ+$) TAR @ 1\% FAR improvements of at least 7.1\% and 1.57\%, respectively. 



\end{abstract}

\section{Introduction}

Over the last decade, facial recognition (FR) systems have significantly advanced in terms of recognition---verification (1:1 matching) and/or identification (1:N matching)---performance. Despite overall FR enhancements across progressively increasing variations in pose, illumination, expression, occlusion, resolution, and scale, FR systems often under-perform~\cite{fondje2020cross, fondje2022learning, sengupta2016frontal}, when matching gallery and query (probe) images with relatively large perspective differences (i.e., greater than 60 degrees). Therefore, FR systems rarely manifest properties of pose invariance, which are ideal for unconstrained recognition and non-cooperative enrollment. Our \textbf{motivation} is to address this critical challenge by leveraging 3D priors---probability distributions representing assumptions about parametric models before incorporating specific observations---to bridge the gap between 2D and 3D representations to improve robustness to larger pose variations.


\begin{figure}[t]
   \centering
   \includegraphics[width=\linewidth]{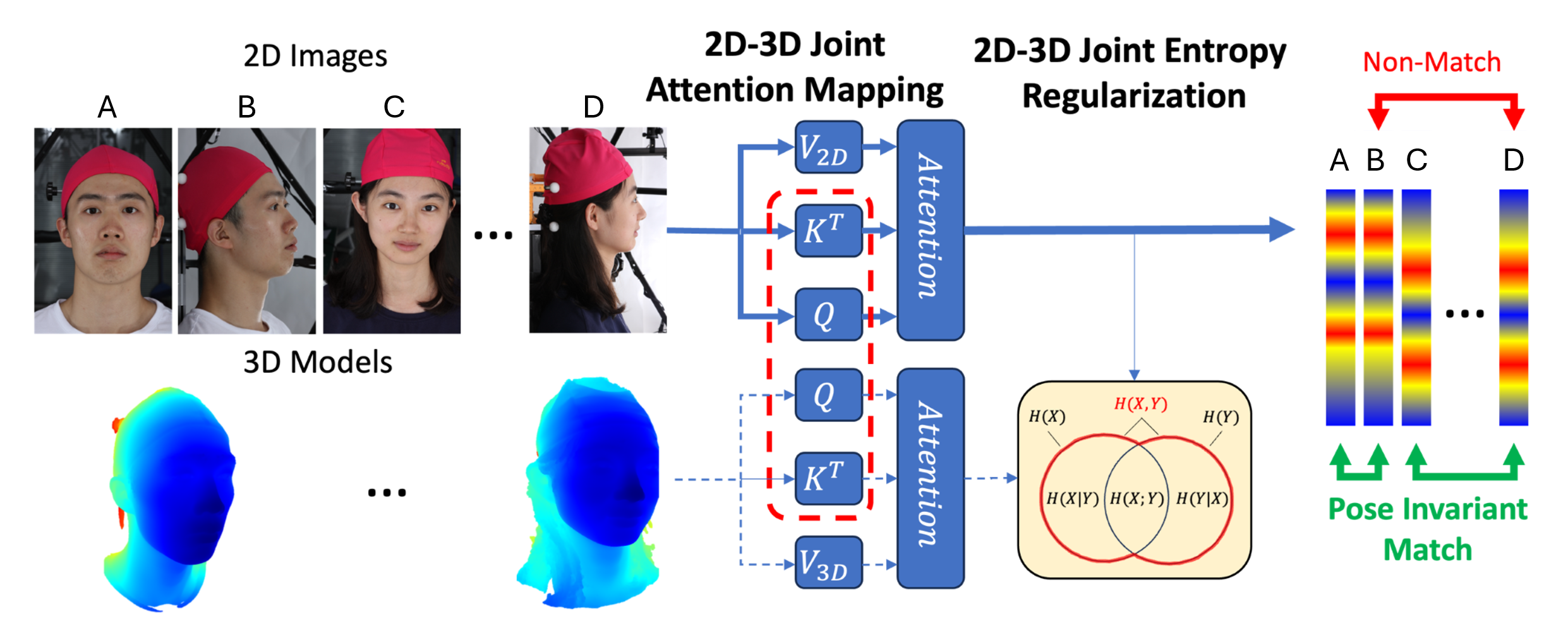}
   \caption{Conventional 2D image embedding networks for FR often exhibit mismatches due to not being invariant to pose. Instead, we introduce a framework that implicitly infers pose invariant 3D information when extracting 2D image embeddings via our proposed 2D-3D Joint Attention Mapping (JAM) and 2D-3D Joint Entropy regularizing loss. This framework processes 2D images (left) to output image representations (right) that are inherently robust to pose variations. For example, images A and B yield similar representations from the same identity, while B and D yield distinct ones despite a similar pose, as they depict different identities. \textit{Subjects consented for publication.}}
   \label{fig:intro}
\end{figure}

Although some FR applications, such as user authentication or airport kiosks, are able to acquire images with optimal frontal viewing angles (pose) and controlled conditions (e.g., lighting and resolution), many government---defense, intelligence, and law enforcement---FR systems operate under less than optimal conditions, such as using off-pose imagery. In operational settings where subjects are not controlled and are non-cooperative, there are significant benefits with addressing the fundamental pose invariance problem for FR. Therefore, the primary objective of this paper is to increase robustness to pose variations by enabling image-based (2D) representations to infer properties of point cloud (3D) representations, which acts as a form of pose invariant guidance (see Figure~\ref{fig:intro}).

Despite advances in unconstrained FR, such as methods that reduce intra-class variance (e.g., Center Loss \cite{wen2016centerloss}, ArcFace \cite{deng2019arcface}), use synthetic views~\cite{masi2016we}, and frontalize faces~\cite{hassner2015effective}, 2D imagery still lacks sufficient understanding of underlying pose invariant 3D information---shape, depth, and texture---due to projective camera geometry. 



To address pose differences, recent literature has employed domain adaptation to model the complex interrelationships among 2D and 3D representations using semantic segmentation~\cite{long2015learning, tzeng2017adversarial} and 3D priors~\cite{hou2021pri3d}. Unlike semantic segmentation methods that directly model relationships among 2D and 3D data,  
our proposed framework \emph{indirectly} establishes consistent representations between domains by aligning and isolating common key discriminative information via 3D priors for 2D image classification to improve robustness to a wide range of pose variations, including profile-to-frontal matching. Similar to Pri3D~\cite{hou2021pri3d}, our method also incorporates 3D priors to benefit 2D image classification. However, instead of directly building 3D priors from rich light detection and ranging (LiDAR) data, our proposed framework enables 2D representation that \emph{indirectly} infer properties of pose invariance from 3D morphable models (3DMMs)~\cite{booth20173DMM} of faces.


Therefore, a new framework is proposed to enhance 2D FR by jointly learning to focus on both spatial and channel information that permits interoperability between 2D and 3D facial analysis. The learned relationships between 2D and 3D representations are enabled by an implicitly learned mapping between the two facial geometries. This framework significantly differs from existing domain adaptation methods that directly adapt RGB images to depth maps \cite{vu2019dada}, which are not pose invariant like point clouds. Additionally, our proposed framework does not rely on the biasing and limiting effects from explicitly masking 3D points that are occluded in corresponding 2D images \cite{zhang2023mx2m}. Our framework employs learned 3D priors from point-clouds, distinctively utilizing them to improve 2D FR capabilities, especially when enrollment (gallery) and query (probe) imagery are acquired from extremely different perspectives. This framework is enabled by a domain adaptive optimization scheme that permits 2D representations during inference without the direct use of 3D input data.


The primary contributions of this work include:
 \begin{compactenum}
   \item a novel \textbf{2D-3D pose invariant FR framework}---a domain adaptation strategy that facilitates pose invariance for 2D images via learned 3D priors,
   \item a new \textbf{2D-3D Joint Attention Mapping (JAM)}---a mapping that uses shared attention to modify the spatial and channel information for both 2D and 3D data,
   \item a new \textbf{2D-3D Joint Entropy (JE) regularization}---a constraint that indirectly promotes 2D and 3D embedding networks to have consistent representations by optimizing the joint attention maps.
 \end{compactenum}

The proposed framework is extensively evaluated across a wide range of pose variations. Not only is the framework evaluated using FaceScape~\cite{yang2020FaceScape}---a dataset containing 2D and 3D data for training purposes---but the trained model is also evaluated for generalization using a large-scale FR dataset (ARL-VTF~\cite{poster2021large}) that only contains 2D images with significant pose variation. Our approach achieves a notable improvement in pose robust performance, surpassing competitive methods with at least 7.1\% and 1.57\% TAR @ 1\% FAR gains for extreme profile views in FaceScape and ARL-VTF, respectively, demonstrating its effectiveness in unconstrained settings.



\section{Relevant Work}

\textbf{2D FR}:
Over the last decade, 2D FR has significantly advanced due to improved tools/resources (e.g., hardware, application programming interfaces, and deep learning) and availability of large-scale datasets.
DeepFace \cite{taigman2014deepface} and VGG-Face \cite{qawaqneh2017deep} used convolutional neural networks (CNNs) to identify faces, which improved performance under progressively more challenging resolution/quality, pose, illumination, and expression conditions. FaceNet \cite{schroff2015facenet} enhanced FR by mapping facial images into a compact Euclidean space. SphereFace \cite{liu2017sphereface} introduced an angular margin loss to improve discrimination. Instead, CosFace \cite{wang2018cosface} optimized the cosine margin, and ArcFace \cite{deng2019arcface} enhanced FR with a tighter angular margin between classes. Pose-invariant methods such as those by Cao et al. \cite{cao2018pose}, Yin et al. \cite{yin2017multi}, Tran et al. \cite{tran2017disentangled}, and Zhao et al. \cite{zhao2018towards} have shown promise, though our study emphasizes newer methodologies for evaluating pose-robustness. AdaFace \cite{kim2022adaface}, a more recent approach, adapts the margin based on sample difficulty, marking an evolution in the state-of-the-art. 


While Generative Adversarial Networks (GANs) \cite{creswell2018generative} and Denoising Diffusion Probabilistic Models (DDPMs) \cite{ho2020denoising} have been shown to be competitive for facial analysis \cite{boutros2023idiff,di2021multi,nair2023ddpm,zhang2017generative}, these methods often are computationally complex and vulnerable to artifacts or hallucination effects. FT-GAN~\cite{zhuang2019ft} addressed pose-invariant FR by transforming faces to frontal views through key points alignment. However, face frontalization gets increasingly difficult as faces exhibit more occlusion due to pose. 
 The identity denoising diffusion probabilistic model (ID3PM)~\cite{kansy2023controllable} generated face images with controlled conditions/attributes. 

Instead, in this paper, a more direct, discriminative model that implicitly leverages 3D information is used to mitigate the issues relating to pose, which avoids the need for keypoint (eyes, nose, and mouth) detection, image frontalization, or overly complex synthesis models.

\textbf{3D FR:}
Early 3D FR efforts focused on leveraging geometric data for reliable identification \cite{lei2014efficient}. Exploring the intricacies of pose and expression variations, Kakadiaris \textit{et al.} \cite{kakadiaris2007three} and Wang \textit{et al.} \cite{wang2010robust} converted 3D facial scans to compact meta data and used difference mappings between aligned 3D faces. Furthermore, Kim \textit{et al.} \cite{kim2017deep} applied transfer learning from 2D images to 3D scans, leveraging a 3D augmentation technique for expression synthesis. 


Recently, 3D FR has shifted from hand-crafted geometric analyses to deep learning based methods.
In \cite{jabberi2023face}, ShapeNets were used for voxel-based representations of 3D shapes. While deep learning has shown promise for 3D FR, many of these method still incorporate different regularization strategies involving geometry. For example, \cite{bagchi2014robust,mpiperis20073} used geodesic techniques to exploit surface geometry.
Moreover, deep learning is more capable of processing raw 3D inputs, assuming there is sufficient data. This idea is supported by Bhople \textit{et al.} \cite{bhople2021point}, who emphasize the direct use of point clouds for capturing facial geometry. 
However, a major challenge in deploying 3D FR is the lack of (1) 3D faces in enrollment databases and (2) the high cost quality 3D acquisition systems. This type of infrastructure change would be both difficult and expensive, given that conventional cameras are cheap and ubiquitous. Li \textit{et al.} \cite{li2022comprehensive} provide a comprehensive review of 3D FR. Instead, our framework uses a 2D-3D domain adaptation approach to be able to leverage beneficial characteristics of 3D faces while retaining the scalability of 2D FR, which partially reduces the burdens identified in 3D FR systems. 

\textbf{2D-3D Domain Adaptation:}
Synergy between 2D imagery and 3D data allows contextual information from 2D images to predict 3D geometries, and vice versa. Therefore, domain adaptation not only considers intra-domain discriminability, but also performs cross-domain matching. 

\textbf{2D Priors to Enhance 3D:} Recently, techniques like joint 3D multi-view prediction networks fuse RGB and geometric data, such as 3DMV \cite{dai20183dmv} which integrates 2D feature maps for enhanced 3D semantic scene segmentation. In \cite{xu2022image2point} pretrained 2D models are used for 3D point-cloud scene understanding. In \cite{genova2021learning} pseudo-labels are derived from 2D semantic segmentations through multi-view fusion. Self-supervised feature learning methods \cite{jing2021self} exploit cross-modality and cross-view correspondences, enabling joint learning of 2D and 3D features.
In \cite{yan20222dpass} 2D Priors Assisted Semantic Segmentation (2DPASS) is introduced to improve point cloud representation by using 2D images with rich appearance information. In \cite{liu2021learning} 
a contrastive pixel-to-point distillation method is used to transfer information from 2D networks to 3D networks. In \cite{wu2023cross} an unsupervised domain adaptation method is introduced for 3D semantic segmentation, which 
helps address domain shifts. In \cite{deng2017amodal} amodal perception of 3D object detection by inferring 3D bounding boxes from 2D bounding boxes in RGB-D imagery is used. To improve point cloud modeling for 3D shape analysis, \cite{yan2022let} explored cross-modal learning between images and point clouds. Lastly, \cite{jaritz2020xmuda} introduced a unsupervised domain adaptation method
to improve performance in both 2D and 3D domains. Unlike these methods, our proposed framework aims to infer pose invariant 3D information to enhance 2D FR.

\textbf{3D Priors to Enhance 2D:} Conversely, 3D information also benefits 2D perception.  Pri3D~\cite{hou2021pri3d} enriched 2D representations by using 3D geometric cues. CrossPoint~\cite{afham2022crosspoint} enhanced discriminability of 2D and 3D models using a self-supervised cross-modal contrastive learning framework. BPNet~\cite{hu2021bidirectional} improved segmentation by using bidirectional projections between 2D and 3D data, and Liu \textit{et al.} \cite{liu20213d} explored a 3D-to-2D distillation approach.



2D-3D domain adaptation methods \cite{ kundu20183d,wang2010robust,zamir2018taskonomy} have largely been proposed to address scene understanding and segmentation tasks. 
PointMBF~\cite{yuan2023pointmbf} utilized unlabeled RGB-D data to improve spatial awareness in images. In \cite{kweon2022joint}, use 2D self-supervision to enhance 3D semantic segmentation. Mask3D~\cite{hou2023mask3d} used 3D priors as part of 2D vision transformers. Similarly, \cite{arsomngern2023learning,nie2023learning} showed how 2D data can infer 3D structures and incorporate CAD-based insights into image analysis. Chen \textit{et al.} \cite{chen2022self} focused on geometric consistency for image representation. SimIPU \cite{li2022simipu} emphasized unsupervised spatial-aware representation learning.

Instead, we uniquely demonstrate that 2D-3D domain adaptation enables 2D image embeddings to exhibit greater degrees of pose invariance.



\section{Motivation}

\begin{figure*}[htb]
\begin{center}
\includegraphics[width=0.90\linewidth]{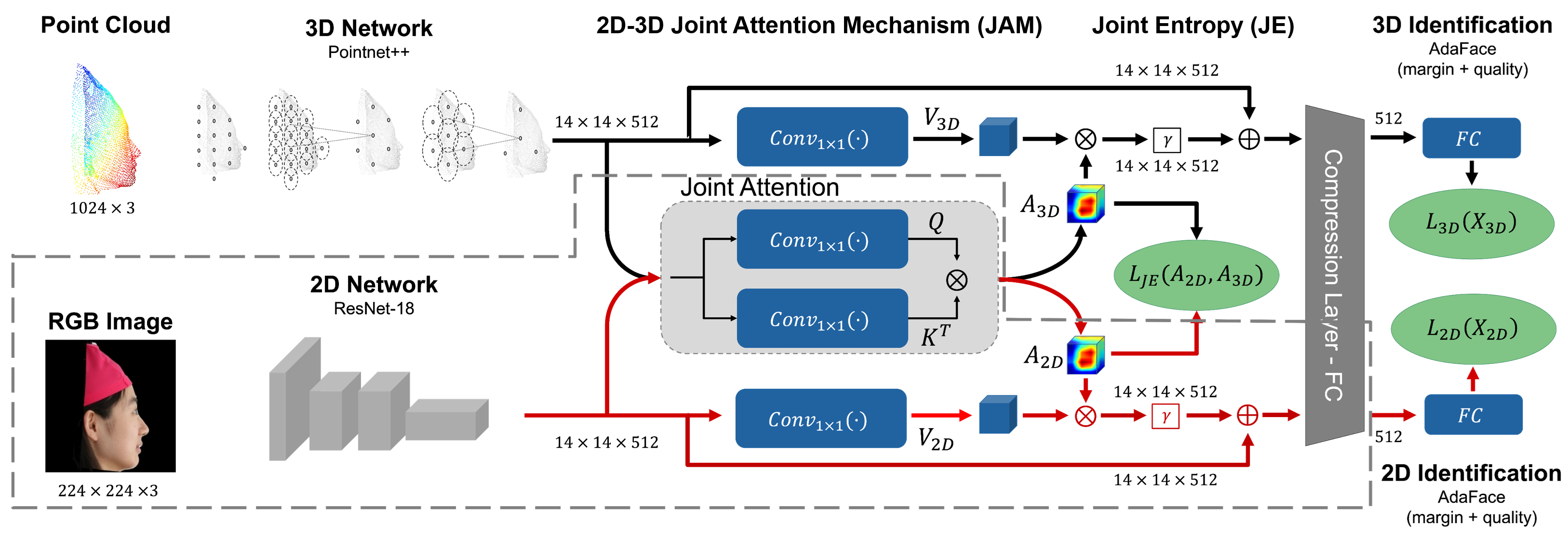}
\end{center}
\vspace{-0.2cm}
\caption[example] 
{ \label{fig:model} 
Our FR framework models pose-invariance by introducing JAM and JE, where this combination promotes consistency of 2D and 3D representations \emph{indirectly} by minimizing the joint entropy over the attention maps ($A_{2D}$ and $A_{3D}$).
While both 2D and 3D data are used during training (\textbf{\textcolor{redarrow}{Red}} and \textbf{\textcolor{black}{Black}} arrows), only 2D images are utilized during inference. \textit{Subjects consented for publication.}}
\end{figure*} 

Pose variations introduce challenges in 2D facial recognition, yet 2D images indeed contain evidence of 3D information (e.g., surface normal and curvature estimates) through priors, such as perspective distortions, shading, and occlusions. Studies in monocular depth estimation \cite{eigen2014depth, godard2017unsupervised} and single-view 3D reconstruction \cite{yan2016perspective, wu2018learning} have demonstrated that deep networks can infer 3D geometry from 2D images, supporting the feasibility of learning 3D priors to enhance 2D feature learning.

Instead of relying solely on monocular depth cues, shape-from-shading~\cite{basri2007photometric} provides a well-established mathematical foundation for inferring 3D structure from 2D images. The shading intensity $I(x,y)$ at a given pixel is a function of surface normals $N(x,y)$ and the light source direction $L$:
\begin{equation}
    I(x,y) = \rho(x,y)\left(N(x,y)\cdot L\right),
\end{equation}
where $\rho(x,y)$ is the surface albedo. This principle suggests that 3D surface geometry can be recovered from 2D images under known or estimated lighting conditions, reinforcing the feasibility of leveraging 2D images to infer 3D priors.
Given this foundation, our method incorporates 3D priors during training to improve pose invariant 2D feature representations. Unlike fusion-based approaches, our framework is designed to operate using only 2D face imagery during inference, but all while exploiting the implicitly inferred 3D information via the proposed offline domain adaptive machine learning framework. Therefore, this design enables better compatibility with existing facial recognition pipelines in comparison to alternative methods that explicitly leverage 3D faces during inference.

The proposed framework is centrally motivated by deriving compatible representations between corresponding 2D and 3D faces to promote pose invariance. Rather than strictly enforcing consistency directly between the 2D and 3D embedding representations to achieve pose invariance---an approach that risks overfitting, as seen in shared attention models that suffer from redundancy due to excessive direct minimization of attention diversity~\cite{li2021your}---our method:
\begin{compactenum}
  \item Establishes dimensional compatibility to ensure that 2D and 3D feature representations can be effectively compared and aligned within a shared latent space.
  \item Formulates a consistent measure for attention in both domains to maintain structured feature correspondences across 2D and 3D embeddings.
  \item Optimizes a probabilistic measure over both attentions to enforce alignment while preserving essential modality-specific variations and preventing overfitting.
\end{compactenum}
By minimizing this measure, namely joint entropy
\begin{equation}
  H(x,y) = -\mathbb{E}[log(P(x,y))],
\end{equation}
our method penalizes differences or inconsistencies between the feature maps, which indirectly promotes consistencies between corresponding 2D and 3D representations themselves. While 2D and 3D data encode modality-specific information, the focus here is on the shared, pose-invariant information that enables better generalization across different poses. Therefore, the proposed methodology aims to boost correlations among 2D and 3D representations to facilitate inferring pose invariant 3D information from 2D data without overfitting.  Although some modality-specific information is sacrificed due to high pose dependency, ideally, remaining information should exhibit (or be able to attain) better robustness to pose. This is not limiting since our model does not depend on or require 3D point clouds during inference, requiring only 2D images.  Therefore, our proposed framework is interoperable with existing facial recognition pipelines.

\section{Methodology}
Our proposed framework is shown in Figure~\ref{fig:model}. This framework includes 2D and 3D backbone architectures to compute discriminative and domain specific representations. The 2D-3D joint attention mapping (JAM) uses query $(Q)$ and key $(K)$ representations across 2D and 3D modalities through a shared parameterization, which forces 2D and 3D attentions to be computed using identical filters. Unlike traditional approaches that share not only attention map parameters but also output values across multiple channels, such as in multi-head channel self-attention model \cite{pecoraro2022local}, our method avoids overfitting by promoting diversity in the learned representations through joint entropy minimization. Then, the attention maps are applied to the 2D value ($V_{2D}$) and 3D value ($V_{3D}$) to emphasize the most discriminative patterns in both domains. The joint entropy regularizing loss, $L_{JE}$, quantifies the consistency between the 2D and 3D attentions, reducing predictive uncertainty and increasing robustness in recognition. The JAM modified representations are compressed and our framework ensures that the compressed representations---encoding JAM modified 2D and 3D embeddings---remain discriminative via domain specific classifiers and losses.

\subsection{Preliminaries}
Given two corresponding facial representations---2D images denoted $X_{2D}$ and 3D point clouds $X_{3D}$---the aim is to learn both consistent and predictable relationships among 2D and 3D embeddings for the purpose of pose invariant 2D FR. Note that $X_{2D}\in\mathbb{R}^{H\times W\times C}$, where $H$, $W$, and $C$ are the image height, width, and number of channels, respectively. $X_{3D} \in\mathbb{R}^{N\times M}$ where $N$ is the number of points in the point cloud and $M$ is the number of dimensions.

\subsection{Backbone Architectures}
Our framework incorporates two distinct state-of-the-art (SOTA) network architectures to process 2D and 3D inputs: Identity Recognition Network 18 (IR-Net-18)\cite{qin2020forward} pretrained using AdaFace \cite{kim2022adaface}, and PointNet++ \cite{qi2017pointnet++}. These networks are denoted by $F_{d}(X_{d};\Theta_{d})$ and are parameterized by $\Theta_{d}$ where $d\in\{2D, 3D\}$. Similar to other domain adaptation methods \cite{fondje2022learning, fondje2020cross}, our framework leverages intermediate representations that are less vulnerable to overfitting to specific domain information. However, due to the differences in structures of the 2D and 3D facial data, our framework is highly asymmetric. We introduce a unique reshaping process to mimic the dimensions of our truncated IR-Net-18. By incorporating an additional convolutional block, we transform the output to a spatial dimension of $(H_{F} \times W_{F} \times C_{F})$, a novel approach for PointNet++ architectures, where $H_{F}$, $W_{F}$, and $C_{F}$ are the feature height, width, and number of channels, respectively. This modification ensures dimensional consistency enabling synergistic training and analysis of facial features.


\subsection{2D-3D Joint Attention Mapping}
\label{sub:jam}
To facilitate pose invariant representations, we introduce a new 2D-3D joint attention mapping (JAM), where attention weights
are shared across both 2D and 3D domains, 
which strategically forces the attention to be computed in the same manner for both 2D and 3D representations. The feature map produced by JAM is mathematically defined as
\begin{equation}
J_d = \gamma\underbrace{softmax\left\{Q(z_d)K^{T}(z_d)\right\}}_{\textrm{Shared Attention}~(A_d)}V_{d}(z_d) + z_d,
\label{eq:JAM}
\end{equation}
where $z_d = F_d(X_d; \Theta_d)$ is the output of the respective backbone network, $Q(\cdot)$ is the shared $1 \times 1$ convolutional layer to compute the query representation, $K^T(\cdot)$ is the shared (and transposed) $1 \times 1$ convolutional layer to compute the key representation, $V_{d}(\cdot)$ denotes the set of $1 \times 1$ convolutional layers used to compute the 2D and 3D value representations, $\gamma$ a learnable parameter influencing attention, and the softmax function
$softmax\left\{a\right\}_{i} = \frac{e^{a_i}}{\sum_{j}e^{a_j}}$
is used to normalize the input to form the attention scores.

The shared attention within our JAM has profound implications---promoting consistency in the joint attention map results in (1) relatively more weight (attention) on spatial and channel information that are discriminative in both 2D and 3D domains and (2) greater similarity between 2D and 3D representations.  
This design ensures a synergistic treatment of spatial information across dimensions, vital for extracting robust, pose-invariant features.


Lastly, the JAM outputs are fed to shared compression and fully connected (FC) layers to reduce dimensionality.

\subsection{2D and 3D Identification}
Our framework leverages AdaFace~\cite{kim2022adaface}, a state-of-the-art and discriminative classifier and adaptive margin based loss function, to ensure that the embedding representations are sufficiently discriminative in terms of independently identifying persons using 2D and/or 3D facial data.

Similar to AdaFace (and other classifiers), the loss functions for 2D and 3D are based on categorical cross-entropy:
\begin{equation}
  L_d = -\mathbb{E}_{y \sim P(Y|X)}\left[\log \big(\hat{P}^{d}(y|x)\big)\right],
\end{equation}
where $P(Y|X)$ is the true distribution of the labels, $Y$, given the input samples, $X$, and $\hat{P}^{d}(y|x)$ is the predicted conditional probability
of the labels given a set of images with modality $d\in\{2D,3D\}$. 

The AdaFace parameters (${m, h, t_a, s}$) were adjusted experimentally from the original settings. Here, $m$ introduces an angular penalty to increase the decision boundary between different classes, encouraging a larger decision boundary. 
 $h$ modulates the additive margin and controls the concentration of the margin scalar prior to clipping, $k$, which correlates to the normalized batch statistics. $t_a$ dynamically modulates the embedding norm's mean and standard deviation by batch, adaptively separating easy from hard samples based on their angular properties. The scaling factor $s$ calibrates the norm of the embedding vector to account for the variable difficulty across samples:
\begin{equation}
\begin{aligned}
  \hat{P}^d(y=y_{i}|x) &= softmax(s (\cos(\theta_{y_i}^{d} + m(1 - hk))\\
  &- m(1 + hk) )),
\end{aligned}
\label{eq:adaface}
\end{equation}
where $\theta_{y_i}^{d}$ denotes the angle between the embedding and the corresponding domain specific classifier weights for label $y_i$. Eq.~\ref{eq:adaface} shows how the model output probability, $\hat{P}^d(y=y_{i}|x)$, for $d \in \{2D,3D\}$ and class specific identity label, $y_i$, integrates the AdaFace angular margin. 
This reformulation ensures that the margin $m$ is factored into the computation of class probabilities, embedding the discriminative method of AdaFace directly into our model's output. This enhances FR by promoting greater angular separation between embeddings of different classes, which improves robustness to limited variations in pose.



\subsection{Joint Entropy Optimization}
Our framework minimizes the joint entropy between the 2D and 3D attention maps:
\begin{equation}
\begin{aligned}
\label{eq:je}
  L_{JE}(A_{2D}, A_{3D}) &= \\
  -\mathbb{E}_{a_{2D},a_{3D} \sim P(A_{2D},A_{3D})}&\left[\log\big(P(a_{2D},a_{3D})\big)\right],
\end{aligned}
\end{equation}
where $\mathbb{E}_{(a_{2D},a_{3D}) \sim P(A_{2D},A_{3D})}[\cdot]$ denotes the expected value over the joint probability distribution of attention maps $A_{2D}$ and $A_{3D}$, 
This probability distribution is calculated based on the discretized attention maps, $\hat{A}_{2D}$ and $\hat{A}_{3D}$, which represent 2D and 3D attention values mapped to discrete bins. After discretization, we compute the histogram over pairs of discretized values $(a_{2D}, a_{3D})$, where $a_{2D}\in\hat{A}_{2D}$ and $a_{3D}\in\hat{A}_{3D}$, and normalize $P(a_{2D}, a_{3D}) =
  \frac{1}{N^2} \sum_{i,j}\mathbb{I}\{A_{2D}(i)\!\!=\!\!a_{2D}, A_{3D}(j)\!\!=\!\!a_{3D}\}$,
where  $N$ is the total number of elements in each attention map $\hat{A}_{2D}$ and $\hat{A}_{3D}$, which both contain the same number of elements.  $\mathbb{I}\{\cdot\}$ is the indicator function. 
This construction ensures that the joint probability distribution is computed across all combinations of discretized values from the 2D and 3D attention maps, then normalized to sum to 1. 

By minimizing Eq.~\ref{eq:je}, our framework helps the model to learn shared attention patterns that reduce inconsistencies between the 2D and 3D modalities. This results in more robust pose-invariant representations, as consistent patterns across both attention maps are emphasized.
The attention mechanisms $A_{2D}$ and $A_{3D}$ are defined according to Eq.~\ref{eq:JAM}, for $d \in \{2D, 3D\}$. The entropy minimization encourages the model to simultaneously learn representations that are informative and discriminative across both modalities.

%
%


Our framework optimizes the following loss function:
\begin{equation}
  L = L_{2D}(X_{2D}) + L_{3D}(X_{3D}) + L_{JE}(A_{2D}, A_{3D}),
\end{equation}
where $L_{2D}(X_{2D})$ is the categorical cross entropy of the 2D pipeline, $L_{3D}(X_{3D})$ is the categorical cross entropy of the 3D pipeline, and $L_{JE}(A_{2D}, A_{3D})$ is the joint entropy loss (Eq.~\ref{eq:je}).  Each loss term was weighted equally, which was determined empirically.  By jointly optimizing our framework using this loss, we are encouraging the model to learn both 2D and 3D representations that are highly discriminative while also promoting consistency between the 2D and 3D attention maps (and indirectly the representations).

In summary, by implementing joint entropy optimization in our 2D-3D mapping, we ensure the synergy of the attention mechanisms, fostering a uniform learning paradigm. This facilitates a model that achieves pose-invariant FR by minimizing entropy across attention maps and establishes a strong foundation for generalized and consistent identification, enhancing robustness and reliability of recognition systems in practical applications.

\vspace{-0.1cm}
\section{Experiments \& Results}
Our framework is rigorously evaluated on diverse datasets to determine the robustness to pose variations and generalization.
Detailed evaluations against benchmarks demonstrate the efficacy of JAM and JE integration in FR tasks across varying poses. We compared to SOTA regularizers such as Maximum Mean Discrepancy (MMD) \cite{gretton2012kernel} and Correlation Alignment (CORAL) \cite{sun2016deep}. We also evaluated over a 2D-3D domain adaptive method based on ResNet-50 \cite{he2016deep} and DGCNN \cite{phan2018dgcnn}, listed as CrossPoint \cite{afham2022crosspoint}, that uses Intra-Model Correspondence (IMID) and Cross-Model Correspondence (CMID) regularizers. In addition, we tested using regularizers from \cite{afham2022crosspoint} on other backbone architectures denoted as IRNet + IMID + CMID that uses PointNet++ for the 3D pose-invariant model.

\subsection{Datasets and Protocols}

\textbf{FaceScape}~\cite{yang2020FaceScape}:  This dataset is a comprehensive collection of facial images and meshes derived from diverse viewpoints under various lighting conditions. This dataset provides 
3DMMs using simultaneously captured images and associated camera parameters. 
We use a subset of $140,000$ samples spanning 140 identities. Each identity has 20 different expressions, from which we sample 50 images per expression.
To evaluate, we randomly split the dataset by identity---forming disjoint training ($70\%$ of identities) and evaluation ($30\%$ of identities)  sets---where the evaluation set contains $\sim$47 identities. The gallery is constructed using 5 samples per identity, and the remaining samples are probe images.


\textbf{ARL-VTF}~\cite{poster2021large}: This dataset provided a secondary evaluation platform: 100 identities, 6 gallery images, 5 probe images, and camera designations to maintain focus on visible spectra, independent of thermal imagery. We use the G\_VB0- listed in their method for gallery and for the probes we use P\_VB0 for frontal probes and P\_VP0 for pose probes. 
Additionally, ARL-VTF does not use segmentation masks to remove background as we have in Facescape.  However, the images have consistent background information.
Models that are trained on FaceScape were used to evaluate performance on ARL-VTF, offering a generalization performance perspective.


This multi-dataset strategy underlines our framework's robustness in handling pose variance and its generalization potential across varying conditions and modalities, towards achieving reliable pose-invariant FR.

\subsection{Performance Metrics}
To mimic operational relevance, we compute the cosine similarity between 2D enrollment and query embedding to produce match scores. Sets of match scores are used to perform Receiver Operating Characteristic (ROC) analysis, including computing True Accept Rate (TAR) @ 1\% False Accept Rate (FAR), Equal Error Rate (EER), and Area Under the Curve (AUC). Specifically, TAR @ 1\% FAR evaluates the effect of pose variations on FaceScape and ARL-VTF. 
 Furthermore, average TAR @ 1\% FAR is evaluated over all the bins and other metrics such as EER and AUC are measured. Testing focused exclusively on 2D pipelines, aligning with our study's discriminative premise.

\subsection{Implementation Details}
\textbf{Preprocessing:}
Our preprocessing for FaceScape aligns input meshes manually to ascertain a canonical frontal pose and crops them to the facial area. Images are scaled to $224 \times 224 \times 3$, with the longest dimension aligned within $[1, 224]$ and the orthogonal dimension cropped to 224 pixels. Pose differentials are calculated from a frontal image using Euler angles. We filter out segmentation masks and normalize each mesh's rotation before sampling about 52 unique samples per expression from the point cloud, which is downsampled to 1024 equidistant points. The 2D and 3D samples were paired by sampling an aligned ID and expression.  The 3D view point was then reprojected from the 3D frontal pose mesh using the camera projection matrix of the 2D image view point.

\textbf{Training:}
Our training protocol was meticulously designed to optimize model performance and ensure robustness across different facial poses. The framework was trained on three folds to enhance generalizability, using a NVIDIA RTX 3090 Ti GPU for efficient computation. Each fold underwent training until convergence was reached, determined by the lack of improvement TAR @ 1\% FAR on the validation set for 9 consecutive epochs, with a minimum threshold of 10 epochs for initial training.

Our AdaFace classifier settings were adjusted heuristically from the recommended settings to $m=0.5, h=0.0, t_a = 0.01, s=64$. We found these parameter values to be best optimized for the subtle differences in the facial features that were highly activated in our dataset. Additionally, these parameters were optimized initially for 2D training only and kept consistent when adapted with the joint 2D-3D attention mechanism training framework.

Aligned 2D and 3D samples, sharing the same pose and identity, ensure consistency in learning pose-invariant features. The models were trained with a batch size of 64, employing Stochastic Gradient Descent (SGD) as the optimizer. Initial learning rate was set at 0.001, with a weight decay of $5e^{-4}$, and learning rate decay milestones were set at epochs 8, 12, and 14 to fine-tune the training process.

Model checkpoints were evaluated based on their performance on the validation set, specifically focusing on achieving the best TAR @ 1\% FAR.  All performance metrics reported are averaged over three independent training cycles, where each model was evaluated on a different population subset. This systematic training and validation approach underpinned the effectiveness of our framework, contributing to its ability to maintain high recognition accuracy across varying poses and conditions.

{\footnotesize
\begin{table*}[!h]
\caption{Overall benchmark and varying pose performance using FaceScape.}
\begin{center}
\resizebox{0.86\textwidth}{!}{
\begin{tabular}{ccp{0.9cm}p{0.9cm}p{0.9cm}p{0.9cm}p{0.9cm}ccc} 
 \hline
 \multirow{1}{*}{Methods} & \multirow{1}{*}{Pose Invariant Source} & \multicolumn{5}{c}{TAR{@}1\%FAR (Pose)} & \multicolumn{3}{c}{Average} \\
  2D Inference and Losses & Used during Training Only & 0-10 & 10-30 & 30-60 & 60-90 & 90+ & TAR{@}1\%FAR & AUC & EER\\ 
 \hline
 CrossPoint~\cite{afham2022crosspoint} & DGCNN & 94.876 & 91.783 & 85.540 & 74.736 & 71.101 & 78.203 & 98.560 & 5.380 \\
 GhostFaceNets~\cite{alansari2023ghostfacenets} + AdaFace~\cite{kim2022adaface} + TS~\cite{hinton2015distilling} & PointNet++ & 87.579 & 82.835 & 75.956 & 66.161 & 63.805 & 69.418 & 97.429 & 7.282 \\
 DREAM~\cite{cao2018pose} + AdaFace~\cite{kim2022adaface} + TS~\cite{hinton2015distilling} & PointNet++ & 91.387 & 87.711 & 79.858 & 67.942 & 62.563 & 70.919 & 98.416 & 5.848 \\
 IRNet~\cite{qin2020forward} + AdaFace~\cite{kim2022adaface} + TS~\cite{hinton2015distilling} & PointNet++ & 90.716 & 87.860 & 82.665 & 65.253 & 62.990 & 71.636 & 98.306 & 6.177 \\
 IRNet~\cite{qin2020forward} + AdaFace~\cite{kim2022adaface} + MMD~\cite{gretton2012kernel} & PointNet++ & 94.240 & 91.723 & 86.568 & 75.250 & 71.503 & 78.315 & 98.275 & 6.314 \\
 IRNet~\cite{qin2020forward} + AdaFace~\cite{kim2022adaface} + CORAL~\cite{sun2016deep} & PointNet++ & 93.562 & 91.062 & 85.241 & 72.852 & 69.363 & 77.723 & 98.233 & 6.373 \\
 IRNet~\cite{qin2020forward} + IMID~\cite{afham2022crosspoint} + CMID~\cite{afham2022crosspoint} & PointNet++ & 94.351 & 91.697 & 86.692 & 75.544 & 71.686 & 79.239 & 98.325 & 6.115 \\
 IRNet~\cite{qin2020forward} + AdaFace~\cite{kim2022adaface} + RST~\cite{fondje2020cross} & PointNet++ & 95.207 & 92.546 & 87.206 & 76.145 & 71.810 & 79.843 & 98.531 & 5.775 \\
 IRNet~\cite{qin2020forward} + AdaFace~\cite{kim2022adaface} + RST~\cite{fondje2020cross} + TS~\cite{hinton2015distilling} & PointNet++ & 93.072 & 89.060 & 81.726 & 67.689 & 62.866 & 72.708 & 97.454 & 7.870 \\
 \hline
 Ours + JAM + JE & PointNet++ & 97.102 & 95.811 & 91.471 & 81.575 & \textbf{78.910} & 84.477 & 98.932 & 4.816\\
 Ours + JAM & PointNet++ & \textbf{97.409} & \textbf{95.946} & \textbf{91.912} & \textbf{82.086} & 78.501 & \textbf{85.206} & \textbf{99.011} & \textbf{4.637}\\
 \hline
\end{tabular}
}
\label{lab:total_bench}
\end{center}
\end{table*}
}
\vspace{-0.1cm}
\subsection{Quantitative Results}
Our experiments on the FaceScape dataset resulted in superior results across all angles, especially large pose differences (Table~\ref{lab:total_bench}). For example, our framework achieves 7.8\% improvement over CrossPoint for profile views. Moreover, the difference between frontal and profile performance is 18.19\% for our method and 23.78\% for CrossPoint. Particularly compelling is the $90^\circ+$ column where Ours + JAM + JE has the highest TAR @ 1\% FAR providing the best pose invariance over other methods. Without the Joint Entropy minimization, Ours + JAM showcased the highest performance (by a marginal amount) on FaceScape across all other poses and average metrics. Despite inconsistencies across poses closer to frontal favoring Ours + JAM, we notice that when JE is used, it regularizes and helps with generalizability on other datasets encouraging overall performance as described by our evaluation on ARL-VTF.  

\begin{figure}[t]
   \centering
   \begin{subfigure}[b]{0.49\linewidth}
     \centering
     \includegraphics[width=\textwidth]{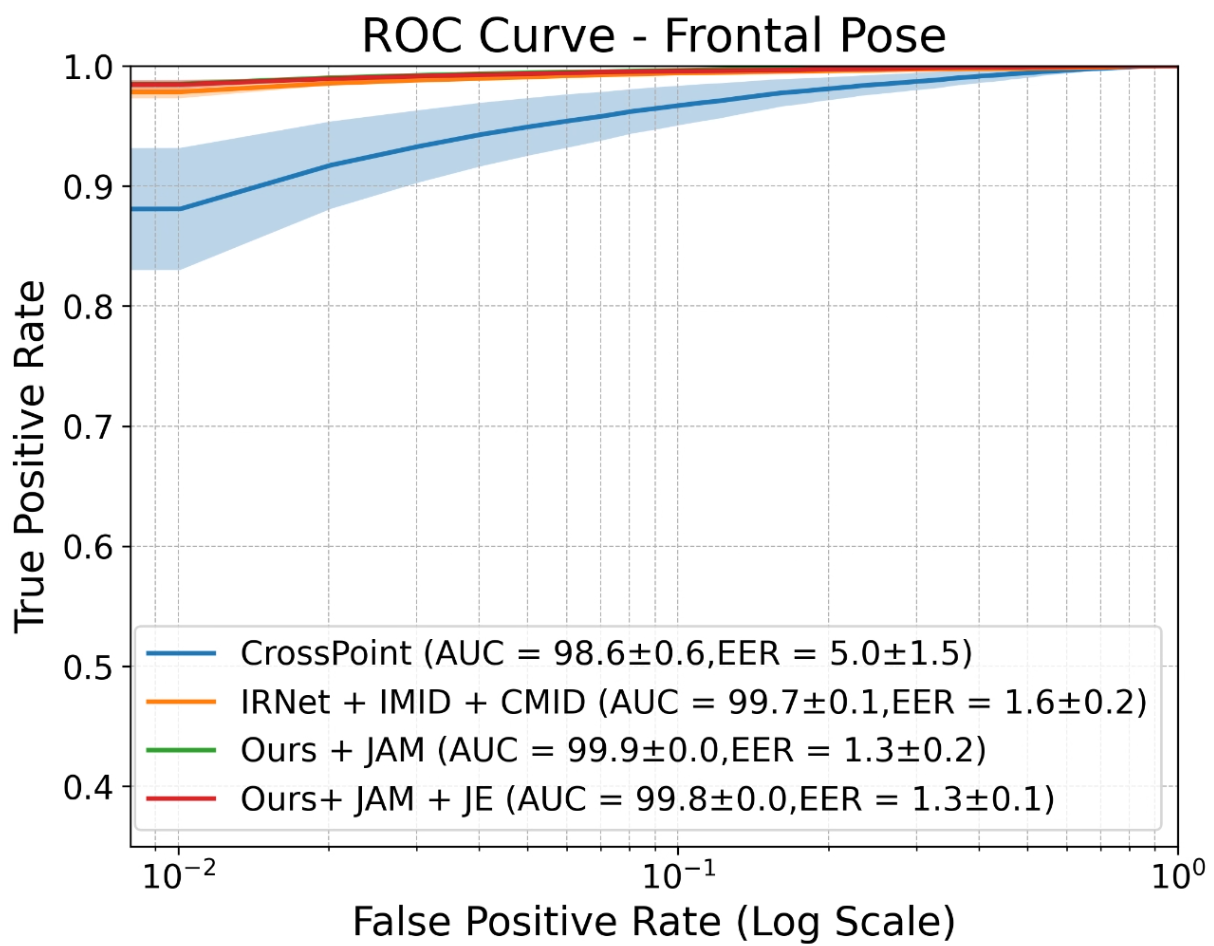}
     \caption{}
     \label{fig:roc_a}
   \end{subfigure}
   \begin{subfigure}[b]{0.49\linewidth}
     \centering
     \includegraphics[width=\textwidth]{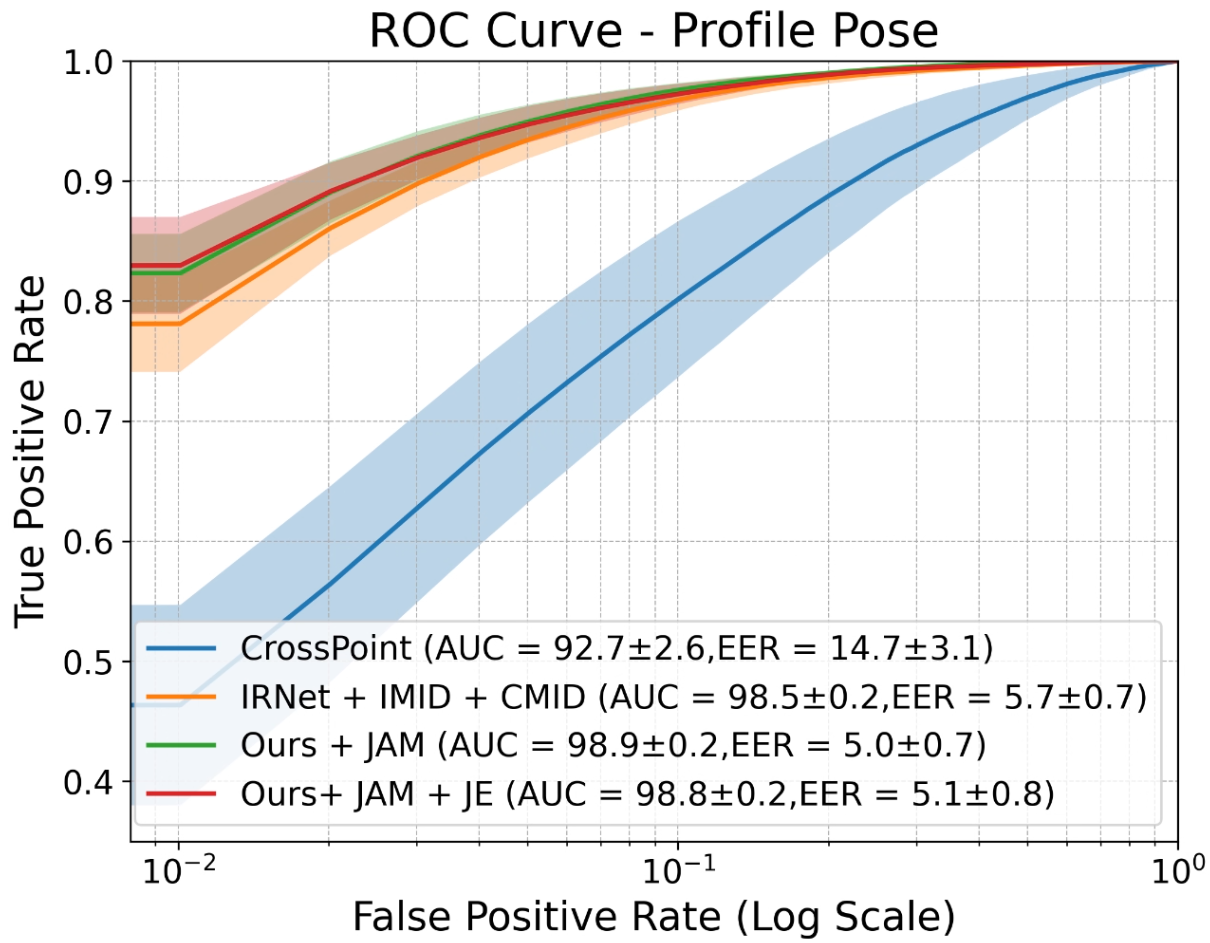}
     \caption{}
     \label{fig:roc_b}
   \end{subfigure}
   \caption{ROC Curves for frontal and profile poses: (a) Frontal pose evaluation of embedding networks with AUC and EER. (b) Profile pose analysis using the same metrics. Shaded regions show variability, representing one standard deviation from the mean.}
   \label{fig:roc}
\end{figure}

Furthermore, the ROC curves in Figure~\ref{fig:roc} show that our framework achieves better pose invariance compared to CrossPoint. In Figure~\ref{fig:roc_a}, Ours + JAM + JE achieved the highest AUC and lowest EER for the frontal condition. Likewise, profile conditions (Figure~\ref{fig:roc_b}), Ours + JAM + JE had the best AUC but with marginally higher EER compared to Ours + JAM, which did not include joint entropy.

The results from the ARL-VTF dataset are shown in Table~\ref{lab:vtf_bench}. Here, Ours + JAM + JE demonstrates superior generalization performance and sensitivity, as indicated by the low EER, on this dataset. In particular, in comparison to Poster \textit{et al.} \cite{poster2021large}, which had the advantage of being trained on ARL-VTF data, our framework (including the use of joint entropy) achieved a 26.128\% improvement under the pose condition but a decrease by 2.605\% for the frontal condition. We suspect the drop in frontal performance is due to a domain shift between FaceScape and ARL-VTF. Nonetheless, our framework achieves the best average performance.

\begin{figure}[t]
   \centering
   \begin{subfigure}[b]{0.4\linewidth}
     \centering
     \includegraphics[width=\textwidth]{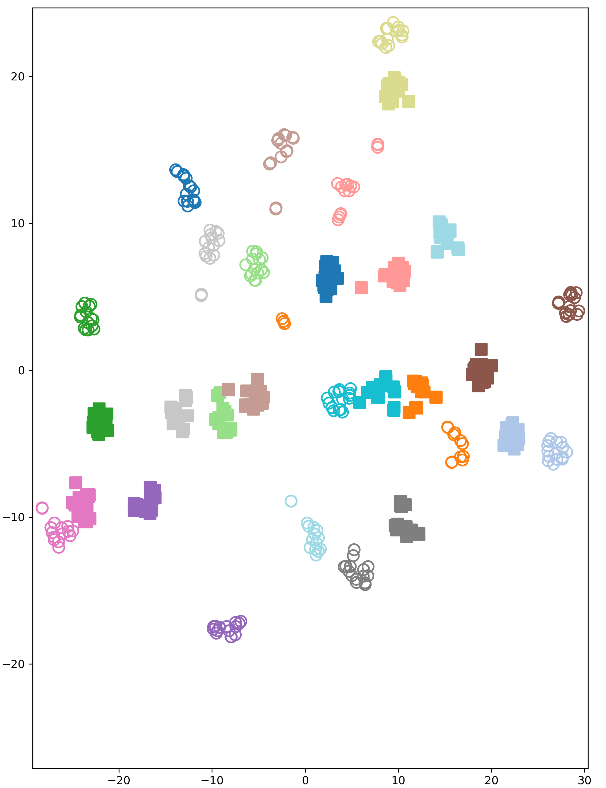}
     \caption{}
     \label{fig:tsne_a}
   \end{subfigure}
   \begin{subfigure}[b]{0.397\linewidth}
     \centering
     \includegraphics[width=\textwidth]{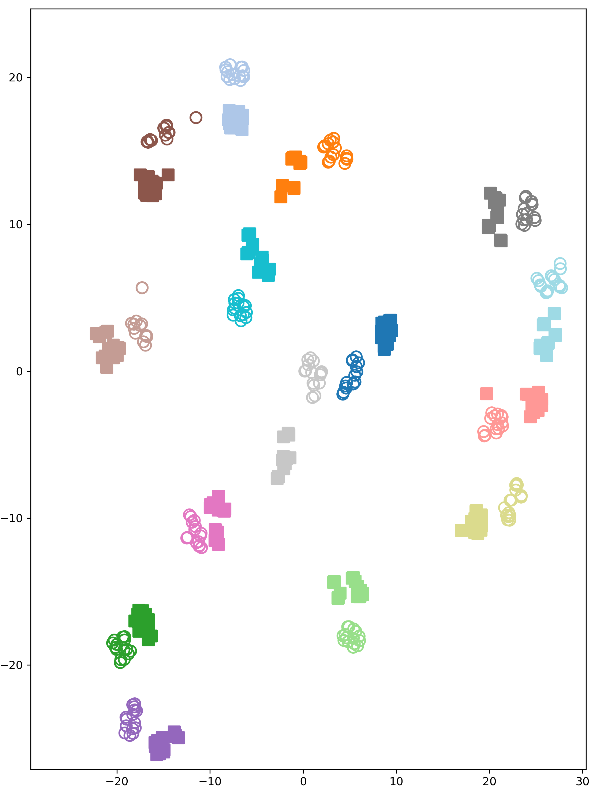}
     \caption{}
     \label{fig:tsne_b}
   \end{subfigure}
   \begin{subfigure}[b]{0.166\linewidth}
    \centering
    \includegraphics[width=\textwidth]{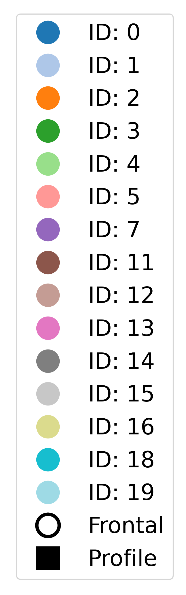}
    \caption{}
   \end{subfigure}
   \caption{t-SNE plots comparing (a) CrossPoint (a) and  Ours + JAM + JE (b). In the legend (c) subjects IDs are indicated by color and poses are indicated by shape (circles for frontal, squares for profile).  Ours + JAM + JE shows better separability across identity and pose.  Notably, IDs: [0, 4, 5, 12, 19] mark improvement with closer convergence of pose embeddings, demonstrating consistency of our method across poses.}
   \vspace{-0.3cm}
   \label{fig:tsne}
\end{figure}
 {\footnotesize
\begin{table*}[h]
\caption{Architecture comparison for 2D and 3D networks to be used for domain adaptation.}
\vspace{-0.25cm}
\begin{center}
\resizebox{0.62\textwidth}{!}{
\begin{tabular}{p{0.001cm}p{0.001cm}cp{0.9cm}p{0.9cm}p{0.9cm}p{0.9cm}p{0.9cm}ccc} 
 \cline{3-11}
 & & \multirow{2}{*}{Methods} & \multicolumn{5}{c}{TAR{@}1\%FAR (Pose)} & \multicolumn{3}{c}{Average} \\
 & & & 0-10 & 10-30 & 30-60 & 60-90 & 90+ & TAR{@}1\%FAR & AUC & EER \\ 
 \cline{3-11}
\parbox[t]{2mm}{\multirow{5}{*}{\rotatebox[origin=c]{90}{2D}}} & \parbox[t]{2mm}{\multirow{5}{*}{\rotatebox[origin=c]{90}{$\overbrace{\hspace{0.75in}}$}}} & VGGNet-16~\cite{simonyan2014very} & 77.469 & 70.779 & 62.441 & 54.295 & 51.639 & 56.068 & 95.324 & 9.468 \\
 & & ResNet-18~\cite{he2016deep} & 89.985 & 85.903 & 80.590 & 71.521 & 67.796 & 72.844 & \textbf{98.667} & \textbf{5.506} \\
 & & ResNet-50~\cite{he2016deep} & 94.173 & 90.988 & 84.577 & 73.012 & \textbf{69.966} & 76.775 & 98.006 & 6.202 \\
 & & DREAM \cite{cao2018pose} & 89.591 & 84.555 & 79.011 & 70.091 & 63.610 & 69.205 & 98.514 & 6.067 \\
 & & GhostFaceNets~\cite{alansari2023ghostfacenets} & 86.411 & 81.691 & 75.203 & 64.445 & 61.022 & 66.204 & 96.983 & 7.954 \\
 & & IRNet-18~\cite{qin2020forward} & \textbf{94.255} & \textbf{91.291} & \textbf{85.915} & \textbf{73.511} & 69.918 & \textbf{78.377} & 98.301 & 6.263 \\
 \cline{3-11}
 \parbox[t]{2mm}{\multirow{3}{*}{\rotatebox[origin=c]{90}{3D}}} & \parbox[t]{2mm}{\multirow{3}{*}{\rotatebox[origin=c]{90}{$\overbrace{\hspace{0.4in}}$}}} & DGCNN~\cite{phan2018dgcnn} & 95.577 & 95.208 & 93.799 & 91.802 & 89.173 & 91.467 & 99.458 & 3.255 \\
 & & PointCNN~\cite{li2018pointcnn} & 52.567 & 51.391 & 49.088 & 46.382 & 45.162 & 45.457 & 95.675 & 11.166 \\
 & & PointNet++~\cite{qi2017pointnet++} & \textbf{96.805} & \textbf{96.355} & \textbf{95.621} & \textbf{95.109} & \textbf{95.040} & \textbf{95.283} & \textbf{99.730} & \textbf{2.252} \\ \cline{3-11}
\end{tabular}
}
\vspace{-0.8cm}
\label{lab:arch_bench}
\end{center}
\end{table*}
}

\begin{figure}[t]
\begin{center}
\includegraphics[width=0.99\linewidth]{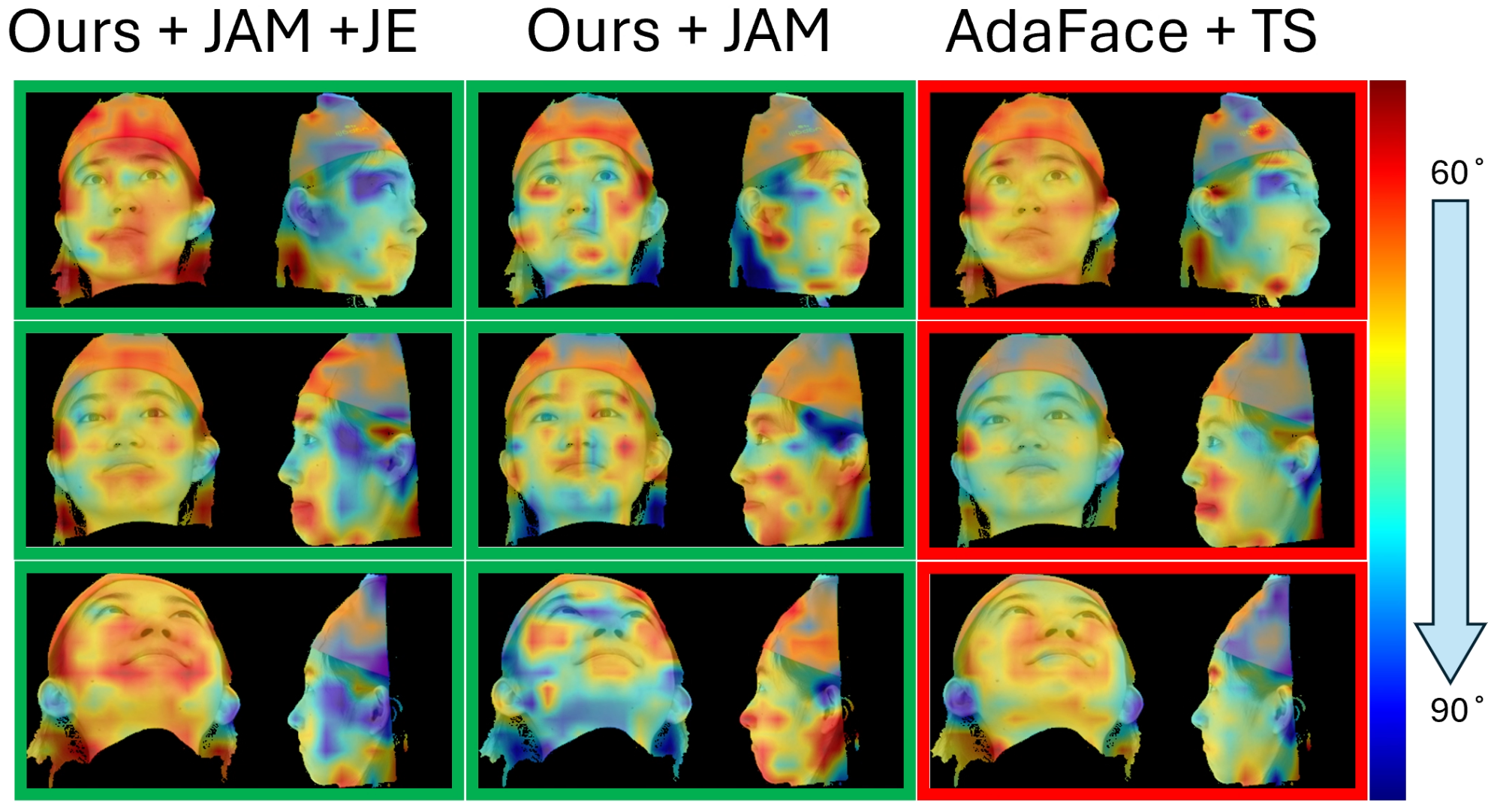}
\end{center}
\vspace{-0.2cm}
\caption[example] 
{ \label{fig:att} Comparison of attention maps superimposed on facial images for frontal and profile pairs across three methods.
\textcolor{darkgreen}{Green} outlines indicate matches above the TAR at 1\% FAR threshold, while \textcolor{red}{Red} shows non-matches. The color bar ranges from red (high attention) to blue (low attention). The arrow shows pose difference from $60^{\circ}$ to $90^{\circ}$ (top to bottom). Ours + JAM + JE exhibits more consistent attention maps across poses and better regularizes attention regions. 
However, Ours + JAM enhances regions of focus, while Ours + JAM + JE de-emphasizes areas prone to misclassification. \textit{Subjects consented for publication.}}
\end{figure} 

\subsection{Qualitative Results}
For qualitative analysis, we compare t-SNE plots and learned attention maps to showcase our framework's robustness to large pase differences. 
Figure~\ref{fig:tsne} shows that the t-SNE visualization for our method with JAM + JE (Figure~\ref{fig:tsne_b}) clusters frontal and profile ($90^\circ+$ pose) faces of the same subject significantly better than CrossPoint (Figure~\ref{fig:tsne_a}). Figure~\ref{fig:att} shows that both our proposed methods more  discriminative than AdaFace + TS, generating attention maps that remain consistent between frontal and profile views. This consistency reflects the robustness of JAM combined with JE minimization, promoting pose-invariant feature representations by regularizing attention across the face. Specifically, the inclusion of JE de-emphasizes regions prone to misclassification with increasing negative attention, which reduces the focus on irrelevant/misleading areas. In contrast, JAM alone shows enhanced attention in discriminative regions, intensifying focus in areas that are key for recognition. However, without JE, this enhancement may overemphasize regions that are inconsistent across large pose differences, leading to less robustness when compared to Ours + JAM + JE. Meanwhile, AdaFace + TS exhibits fewer matches and higher sensitivity to pose variations, with attention maps that are more neutral toward discriminative features in off-pose scenarios. 
Therefore, JE regularizes the JAM based image representations.
\vspace{-0.1cm}
\subsection{Ablation Studies}
To determine the best 2D and 3D backbones for our method on FaceScape, we evaluated each model without domain adaptive regularizers or transforms for identification. For all the methods, besides CrossPoint, we use PointNet++ for our pose-invariant source due to its high performance, as it dominated performance for 3D FR for FaceScape, versus other methods on our 3D benchmark. For our 2D benchmark, we found AdaFace \cite{kim2022adaface} with IRNet-18 backbone to be the highest performing model across most results, with a very small minutia decrease in $90^\circ+$ accuracy against ResNet-50. However, IRNet-18 with AdaFace proved to be a better candidate for domain adaptation to 3D as it would not overfit features to classify in 2D domain lending less parameters to adapted features. Given these results, as shown in Table~\ref{lab:arch_bench}, we chose PointNet++ and IRNet-18 for our 2D and 3D backbone architectures, respectively.

First, we compared using the spatial attention map without JE minimization, as shown by Ours + JAM in Table~\ref{lab:total_bench} to evaluate the effect of implementing joint entropy minimization. The only change was whether or not we minimized with the loss function. Joint entropy minimization improved pose invariance, with additional improvements across all viewing angles when evaluated on another dataset, further highlighting the importance of joint entropy minimization.

We also compared our method to an RST-inspired transformation, as shown in Table~\ref{lab:total_bench} as IRNet + RST. This transformation modulates and optimizes the feature space to 512, which enhances model performance. 

Furthermore, we compared results with teacher-student (TS) networks as in \cite{hinton2015distilling}. In this case, a pretrained 3D classifer acts as the teacher for 2D (student) representations. 
Two methods in Table~\ref{lab:total_bench} incorporated TS this type of training. IRNet-18 + TS, which is uses our 2D network directly fed into the compression layer followed by the 3D teacher, and IRNet-18 + RST + TS, which used an RST~\cite{fondje2020cross} inspired transform as an competitive domain adaptation method.
\vspace{-0.1cm}
{\footnotesize
\begin{table}[!h]
\caption{Benchmark performance and pose analysis for on ARL-VTF (visible only) \emph{when trained on FaceScape.}}
\vspace{-0.25cm}
\begin{center}
\resizebox{\columnwidth}{!}{
\begin{tabular}{cp{0.8cm}p{0.8cm}ccc} 
 \hline
 \multirow{2}{*}{Methods} & \multicolumn{2}{c}{TAR{@}1\%FAR} & \multicolumn{3}{c}{Average} \\
 & Pose & Frontal & TAR{@}1\%FAR & AUC & EER \\ 
 \hline
 CrossPoint~\cite{afham2022crosspoint} & 53.107 & 72.111 & 63.854 & 87.708 & 18.275 \\
 Poster \textit{et al.} \cite{poster2021large} & 28.540 & \textbf{99.790}\footnote{The method presented in Poster \textit{et al.}~\cite{poster2021large} is particularly optimized for frontal face recognition, focusing primarily on enhancing performance under this specific condition.}& 64.165 & 87.875 & 16.265 \\
 DREAM~\cite{cao2018pose} + AdaFace~\cite{kim2022adaface} & 53.609 & 69.796 & 62.771 & 87.151 & 18.323\\
 GhostFaceNets~\cite{alansari2023ghostfacenets} + AdaFace~\cite{kim2022adaface} & 48.602 & 70.481 & 60.987 & 84.709 & 20.847 \\
 IRNet~\cite{qin2020forward} + AdaFace~\cite{kim2022adaface} + MMD~\cite{gretton2012kernel} & 52.521 & 75.074 & 65.287 & 91.478 & 14.321 \\
 IRNet~\cite{qin2020forward} + AdaFace~\cite{kim2022adaface} + CORAL~\cite{sun2016deep} & 52.421 & 75.666 & 65.579 & 91.361 & 14.445 \\
 IRNet~\cite{qin2020forward} + AdaFace~\cite{kim2022adaface} + IMID~\cite{afham2022crosspoint} + CMID~\cite{afham2022crosspoint} & 52.645 & 75.851 & 65.781 & 91.356 & 14.547 \\
 \hline
 Ours + JAM + JE & \textbf{54.678} & 79.185 & \textbf{68.550} & \textbf{93.432} & \textbf{11.594} \\
 Ours + JAM & 53.191 & 76.444 & 66.353 & 91.597 & 14.005 \\
 \hline
\end{tabular}
}
\label{lab:vtf_bench}
\end{center}
\end{table}
}
\vspace{-0.3cm}

\footnotetext[1]{The method presented in Poster \textit{et al.}~\cite{poster2021large} overfits to frontal conditions, yield poor results with pose variations}



\section{Conclusion}

Our proposed methodology addresses the fundamental challenge of performance degradation in 2D pose-variant FR by leveraging 3D priors during training while ensuring a purely 2D inference pipeline. This allows for enhanced robustness without requiring any 3D data at deployment, making our approach potentially scalable for real-world applications. Our primary contributions include the development of a 2D-3D pose invariant domain adaptive framework, a 2D-3D joint attention mapping, and a 2D-3D joint entropy regularizing loss. The limitation of this method is requiring an aligned 2D and 3D dataset for training, despite not requiring 3D data during inference, the requirement for more robust 2D-3D facial datasets is core to our proposed method.


Our methods demonstrated superior performance, especially with extreme pose difference 
and demonstrated robust generalizability and adaptability across datasets, exemplified by the cross-dataset validation with ARL-VTF dataset. Notably, our studies further validated the success of the joint entropy minimization approach,  showing that our framework alleviates FR issues concerning pose differences. 

\textit{The views and conclusions contained herein are those of the authors and should not be interpreted as necessarily representing the official policies, either expressed or implied, of DEVCOM ARL or the U.S. Government.}




\section*{ETHICAL IMPACT STATEMENT}
This work investigates the improvement of pose-invariant facial recognition using publicly available datasets, ARL-VTF and FaceScape. The ARL-VTF dataset was collected under an IRB-approved protocol, ensuring informed consent from participants and additionally giving subjects the ability to opt out from having their imagery appear in publications or presentations. The FaceScape dataset restricts the use of non-publishable subject data, providing a curated list of IDs that are permitted for publication, which demonstrates an attempt to control ethical considerations. While both datasets are widely used in academic research, high-fidelity identifiable data (such as 2D and 3D facial data) carries risks related to re-identification, loss of privacy, spoofing, and other unintended consequences. However, these datasets do implement rather restrictive protocols that help mitigate these risks.

A key concern is the potential for misuse in surveillance and artificial synthesis. While this work is solely focused on scientific advancements in biometric recognition, the broader implications of facial recognition technology should be acknowledged. To mitigate risks, this research does not reconstruct or expose identifiable faces, and any publicly shared models or findings will follow ethical AI principles to prevent misuse.

Additionally, dataset bias must be considered. The FaceScape dataset consists primarily of Asian subjects, while ARL-VTF's demographic diversity is not explicitly documented. However, personal review suggests a diverse subject pool. Although this study does not focus on deployment, fairness in facial recognition remains an important issue, and future research should explore methods that ensure robust performance across demographic groups.

Overall, this research is conducted with a strong awareness of ethical risks and a commitment to responsible AI development. By maintaining transparency in methodology and focusing on scientific contributions rather than direct deployment, this work seeks to advance facial recognition techniques while minimizing potential harms.

{\small
\bibliographystyle{ieee}
\bibliography{egbib}
}

\end{document}